# Geographical hotspot prediction based on point cloud-voxel-community partition clustering

Yan Tang*

College of Software, Northeastern University, Chuangxin Road, 110819, Shenyang, Liaoning, China
*Corresponding author's email: tangy7025@163.com

## ABSTRACT

Existing solutions to the hotspot prediction problem in the field of geographic information remain at a relatively preliminary stage. This study presents a novel approach for detecting and predicting geographical hotspots, utilizing point cloud-voxel-community partition clustering. By analyzing high-dimensional data, we represent spatial information through point clouds, which are then subdivided into multiple voxels to enhance analytical efficiency. Our method identifies spatial voxels with similar characteristics through community partitioning, thereby revealing underlying patterns in hotspot distributions. Experimental results indicate that when applied to a dataset of archaeological sites in Turkey, our approach achieves a 19.31% increase in processing speed, with an accuracy loss of merely 6%, outperforming traditional clustering methods. This method not only provides a fresh perspective for hotspot prediction but also serves as an effective tool for high-dimensional data analysis.

## 1. INTRODUCTION

In the field of Geographic Information Systems (GIS), predicting hotspots entails examining and modeling data to detect and forecast the emergence of events, phenomena, or activities that occur with high frequency or intensity in designated areas. Previous research has often employed this analytical approach to identify and anticipate various incidents, including criminal activities [1,2], natural disasters [3], and biological distributions [4,5].

A common strategy for tackling the challenge of hotspot prediction is clustering [1], with earlier studies [6] also investigating innovative clustering techniques to improve this analytical process. This paper presents a novel clustering methodology: point cloud-voxel-community partition clustering.

Point clouds serve as a prevalent representation of spatial data in three-dimensional domains [8], comprising numerous spatial points, each containing coordinate information (x, y, z) and additional attributes such as color and intensity. When processing spatial data, point clouds offer a detailed and intuitive three-dimensional structure that helps researchers better comprehend and analyze intricate geographical phenomena. Compared to traditional rasterization methods, point cloud approaches can more accurately depict the features of objects within their data dimensions. The point cloud-voxel method builds upon point clouds by subdividing the data into multiple voxels, or three-dimensional grid units, facilitating more efficient spatial analysis and computation, especially suited for large-scale and high-dimensional data processing.

Building upon this foundation, we introduce the concept of communities. Community partitioning [9] is a crucial method in network analysis, commonly employed to identify the close relationships or interactions among nodes. Within the context of GIS, community partitioning can group spatial points (for instance, geographical features represented by point clouds) into units possessing similar characteristics or behaviors. Consequently, different communities may exhibit distinct distributions or trends of hotspots, enabling precise recognition of significant spatial patterns. Notably, our method diverges from traditional practices by utilizing voxel units as the basis for community partitioning rather than conventional data points.

The framework employing point cloud-voxel-community methods was initially applied to the anomaly traffic detection problem [7], facilitating the rapid identification of anomalous clusters and capturing unusual patterns from large-scale data. However, primarily [7] focused on detecting anomalies from a distribution density perspective without considering the data itself. In this paper, we propose a method for geographic hotspot prediction based on point cloud-voxel-community partition clustering: first, we utilize the point cloud-voxel-community framework to conduct foundational partitioning of a significant amount of high-dimensional data points, maximizing efficiency. Subsequently, we perform clustering within the partitioned voxel communities to identify target sets for hotspot predictions. Our method is designed to swiftly capture the underlying pattern information from vast datasets while maintaining accuracy comparable to existing hotspot prediction techniques.

## 2. CONSTRUCTION PROCESS OF POINT CLOUD, VOXEL, AND COMMUNITY IN DATA FEATURE DIMENSIONS

### 2.1 Point cloud mapping

The term "point cloud" is typically defined as a collection of data points P in three-dimensional space $\{P_1, P_2 ... P_n\}$. However, in this paper, we redefine it as the mapping of data distribution across its feature dimensions $P \to p$. This definition extends the conventional understanding of point clouds, moving beyond a mere combination of geometric coordinates to emphasize the information encapsulated by the data points and their structural and relational configurations within a multidimensional feature space.

$$\{P_1, P_2 ... P_n\} \to \{p_1, p_2 ... p_n\} \quad (1)$$

In this extended definition, a point cloud is not solely a collection of spatial coordinates; it also reflects the distribution of intrinsic features and attributes of the data. Each point in the mapped coordinates $p$ intuitively illustrates the correlation of multiple dimensions of representation, which may include features $f$ such as color, texture, reflectance intensity, and temporal data.

$$p_i \to \{f_{1i}, f_{2i} ... f_{ki}\} \quad (2)$$

Thus, point clouds can be viewed as a representation of high-dimensional data, where the distribution in feature dimensions offers insights that are yet to be fully explored. This perspective not only enriches the theoretical understanding of point clouds but also provides a deeper comprehension for data analysis in practical applications.

### 2.2 High-dimensional voxels and community formation

Similar to the previous section, the concept of voxels has been expanded in our discussion. Traditionally, a voxel, or volume element, serves as the fundamental unit for representing three-dimensional data structures. In this context, a voxel is essentially equivalent to a three-dimensional pixel, functioning as a discrete unit that stores spatial information within a three-dimensional grid. However, this concept becomes significantly more relevant in the realm of high-dimensional data analysis. High-dimensional voxels refer to an extension of the basic voxel structure into higher dimensions, which facilitates the representation of complex datasets that transcend conventional three-dimensional boundaries. This evolution is particularly pertinent in fields such as medical imaging and computational biology, where biological phenomena and spatial representations often encompass multiple parameters. By incorporating additional dimensions, high-dimensional voxels enable a richer data representation, capturing not only spatial relationships but also reflecting various intrinsic attribute associations. We numerically normalize the non-discrete features within the point cloud and impose precision limitations, thereby enhancing computational speed. Building upon this, we define a voxel $V$ as a region in high-dimensional space where each dimensional feature of the data is constrained to a specific value range.

$$\begin{cases} \vec{\alpha} = [\alpha_j], & \alpha_j = \min(f_{1j}, ..., f_{ij}, ..., f_{Nj}), \\ \vec{\beta} = [\beta_j], & \beta_j = \max(f_{1j}, ..., f_{ij}, ..., f_{Nj}), \\ 1 \leq j \leq M, & 1 \leq i \leq N, \end{cases} \quad (3)$$

$$\vec{p}_i = [f'_{i1}, ..., f'_{ij}, ..., f'_{iM}]^T, \quad f'_{ij} = \frac{p_{ij} - \alpha_j}{\beta_j - \alpha_j}. \quad (4)$$

Building upon the existing foundations, we have expanded and defined the concept of high-dimensional voxels. Traditionally, a voxel serves as the fundamental unit of three-dimensional data structures, akin to a pixel in three-dimensional space, which holds spatial information. However, the concept of voxels becomes increasingly crucial in high-dimensional data analysis. High-dimensional voxels represent an extension of basic voxel structures to higher dimensions, allowing them to encapsulate complex datasets that transcend conventional three-dimensional boundaries.

In this context, we define a high-dimensional voxel V as a region expressed mathematically by:

$$V = \{\mathbf{x} \in \mathrm{R}^M \mid x_j \in [\alpha_j, \beta_j], \forall j \in \{1,2,\dots,M\}\}. \tag{5}$$

Where $V$ denotes a voxel in high-dimensional space, $x$ is a point in this space, $x_j$ represents the j-th coordinate of $x$, $\alpha_j$ serves as the lower bound for the j-th dimensional feature, determined during the data normalization process, while $\beta_j$ is the upper bound for the same feature, also derived from the normalization process.

To further enhance flexibility in describing these voxels, we introduce two additional symbols: $C_j$ which represents the central point of the j-th dimension, and $R_j$ which indicates the range or extent of that dimension. Incorporating these new symbols, we revise the voxel definition to:

$$V = \{\mathbf{x} \in \mathrm{R}^M \mid x_j \in [C_j - R_j, C_j + R_j], \forall j \in \{1,2,\dots,M\}\}. \tag{6}$$

This definition indicates that the high-dimensional voxel V encompasses a multi-dimensional region, where the characteristic values for each dimension are determined by the center $C_j$ and the range $R_j$.

The construction of our communities is further developed on the foundation of high-dimensional voxels. Specifically, we utilize the Manhattan distance as the basis for community formation. This distance metric focuses on calculating the absolute differences between data points in high-dimensional space to assess their similarity. It is notable for its significantly lower computational complexity compared to Euclidean distance, while also allowing for a more holistic view of the data compared to Chebyshev distance. This Manhattan distance-based approach to community formation enables us to effectively aggregate data points that exhibit similar underlying patterns across multiple feature dimensions, thereby creating more representative and structured communities.

The community C is defined as the set comprising all data points $\mathbf{x}^{(i)}$ that have a Manhattan distance D to at least one other data point $\mathbf{x}^{(j)}$ that is less than a predefined threshold $\epsilon$. Mathematically, this can be expressed as:

$$C = \{\mathbf{x}^{(i)} \in \{\mathbf{x}^{(1)}, \mathbf{x}^{(2)}, \dots, \mathbf{x}^{(N)}\} \mid D(\mathbf{x}^{(i)}, \mathbf{x}^{(j)}) < \epsilon, \forall \mathbf{x}^{(j)} \in \{\mathbf{x}^{(1)}, \mathbf{x}^{(2)}, \dots, \mathbf{x}^{(N)}\}\}. \tag{7}$$

Under this definition, the centroid $C_{\text{centroid}}$ of the community is formulated as follows:

$$C_{\text{centroid}} = \frac{1}{|C|} \sum_{\mathbf{x}^{(i)} \in C} \mathbf{x}^{(i)}. \tag{8}$$

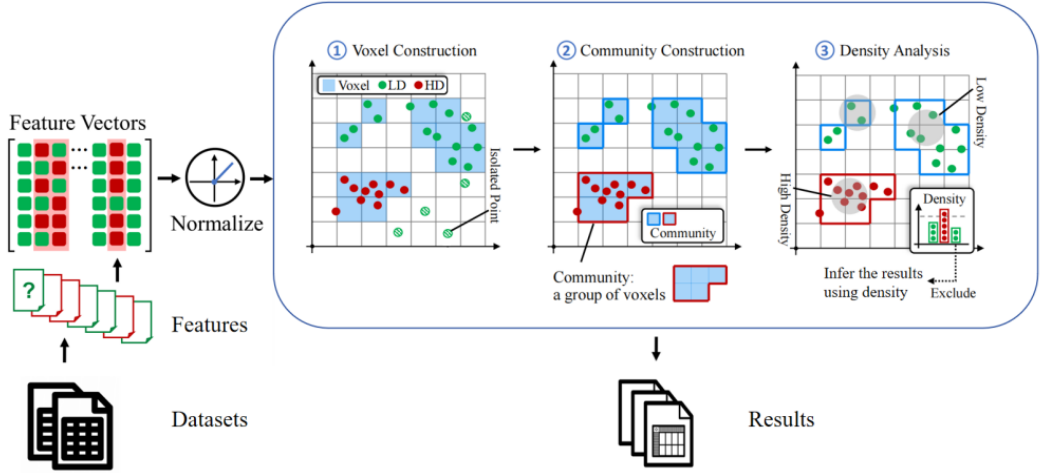

Figure 1. Schematic diagram of the point cloud-voxel-community construction process for high-dimensional data.

## 3. PARTITION CLUSTERING

In the previous phase of the point cloud-voxel-community analysis, we successfully constructed a series of communities based on high-dimensional voxels. Constructing the process is illustrated in Figure 1. These communities are not merely simple aggregates of high-dimensional data; they encapsulate some latent relationships and implicit features inherent in the spatial distribution of high-dimensional data. We posit that the data points within these communities exhibit profound interconnections, reflecting the underlying structure and patterns of data in high-dimensional space.

Building on this foundation, we will further investigate methods for partitioning and clustering. Within each high-dimensional voxel community, we assess the normalized hotspot occurrence probability based on density. Using global statistical parameters, we calculate the expected number of hotspots for a specific community, and we perform k-means clustering within that community, using the expected hotspot count as a parameter. The resulting cluster centers are then adjusted according to the distribution of voxel centers within the community, and a regularization term is applied to yield predicted hotspots.

The adjustment of the cluster center based on the distribution of voxel centers can be expressed mathematically as:

$$\mu_k' = \frac{1}{|C_k|}\sum_{x_i \in C_k} x_i + \lambda \cdot c_k. \tag{9}$$

In this equation, $\mu_k'$ represents the adjusted cluster center for cluster $k$, where $|C_k|$ denotes the number of data points within cluster $C_k$, and $x_i$ are the individual data points in $C_k$. The term $\lambda \cdot c_k$ incorporates a regularization factor $\lambda$ multiplied by the vector $c_k$, which signifies the center of the corresponding community. This adjustment ensures that the cluster centers align more closely with the spatial distribution characteristics of the underlying data.

To ensure the stability of the model and prevent overfitting, a regularization term $R$ can be introduced to balance the distribution of the cluster centers:

$$R = \sum_{k=1}^{K} \| \mu_k - c \|^2. \tag{10}$$

where $c$ represents the mean of the community.

This approach aims to subdivide the dataset within the same community into multiple non-overlapping subsets, where data points in the same subset exhibit high similarity, while the similarity between different subsets is relatively low. We anticipate that by integrating the methodologies from the previous two sections, we can delve deeper into the intrinsic and structural characteristics of the data, thereby uncovering valuable information within the dataset.

## 4. EXPERIMENTAL RESULTS

In our research, we utilized a unique dataset that chronicles the archaeological sites of Turkey, with a particular emphasis on the Iron Age. This comprehensive dataset synthesizes information from the Archaeological Settlements of Turkey (TAY) Project[10] with geospatial data derived from OpenStreetMap, thereby offering a nuanced portrayal of Turkey's historical and cultural evolution during this pivotal epoch. We believe this dataset possesses significant exploratory potential and exemplifies the efficacy of our methods in addressing hotspot prediction challenges.

### 4.1 Performance and comparison

We executed over a hundred iterations of K-MEANS, C-MEANS[1], and our proprietary algorithm on this dataset. After excluding outlier runtimes, we randomly selected 100 data points from the runtime collections of each algorithm for comparative analysis. The results are shown in Figure 2:

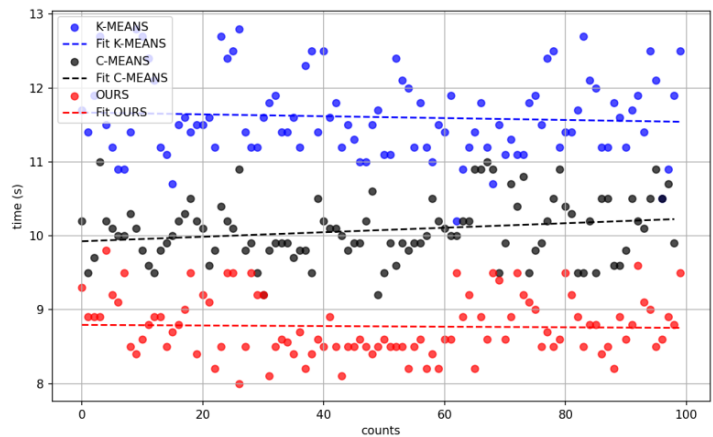

Figure 2. Comparison of running times for K-MEANS, C-MEANS[1] and our method after removing outliers based on 100 random trials.

We selected an observable subset from the original dataset and applied our method to it, as shown in Figure 3. Our objective is to demonstrate that our approach can effectively perform point cloud-to-voxel-community steps in high-dimensional space and successfully capture commonalities within the high-dimensional representations through partition clustering to accomplish hotspot prediction. The results are shown in Figure 4.

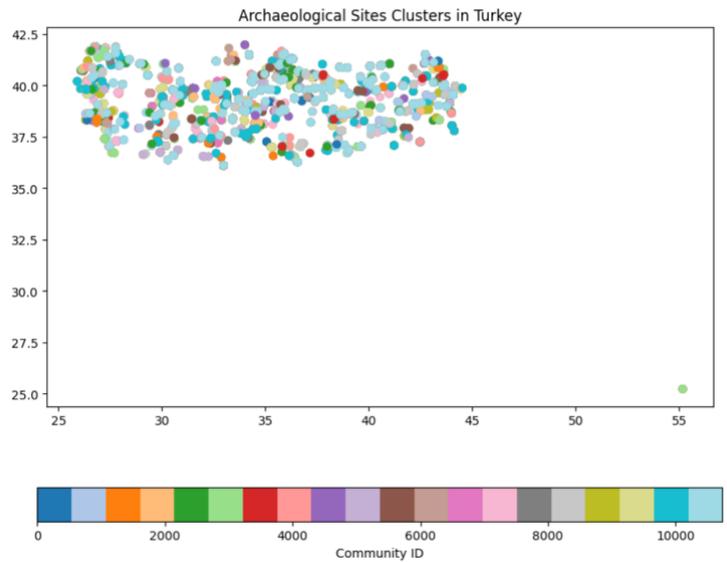

Figure 3. A running example of our method on a subset of the TAY dataset, which shows the construction process of different voxels and communities.

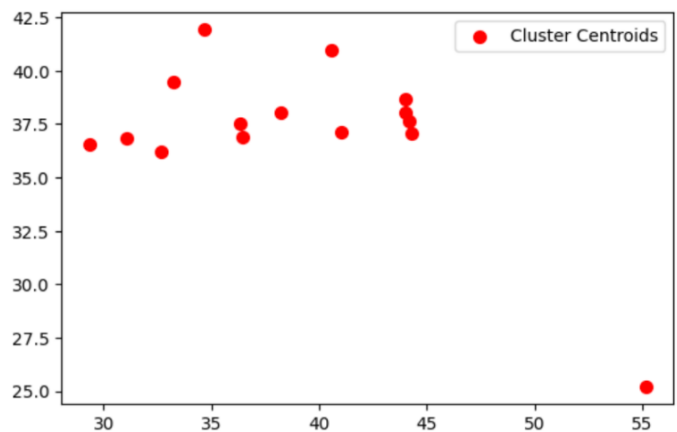

Figure 4. Hotspot prediction results of our method on a subset of the TAY dataset.

Due to the inherent uncertainty associated with the optimal solutions to prediction problems, we randomly removed certain points from the original dataset and employed the aforementioned three methods to complete hotspot prediction. We measured performance by compiling the predicted points that were distributed within the same region as the removed points. The data in Table 1 shows that our method achieved a significant speed improvement of 19.31%, while the accuracy loss was only 6% compared to optimized methods, outperforming basic clustering approaches. These results demonstrate the effectiveness of our method in rapidly and accurately addressing hotspot prediction challenges.

Table 1. Comparisons of Average Time Spent(s) and accuracy (%) in the TAY dataset.

| Method | Average Time Spent(s) | Accuracy |
| --- | --- | --- |
| K-MEANS | 11.61 | 42/66(63.63%) |
| C-MEANS | 10.08 | 57/66(86.36%) |
| OURS | **8.75** | 53/66(80.30%) |

## 5. CONCLUSION

In conclusion, this study introduces an innovative approach to geographical hotspot prediction through the implementation of point cloud-voxel-community partition clustering. Addressing the limitations of existing solutions in the field of geographic information, our method enhances analytical efficiency by leveraging high-dimensional data representations. The experimental results, derived from a comprehensive dataset of archaeological sites in Turkey, demonstrate a significant improvement in processing speed by 19.31%, with a minimal accuracy loss of only 6% compared to traditional clustering techniques. This research not only contributes a new perspective for hotspot prediction but also establishes a robust framework for analyzing complex high-dimensional datasets. The findings underscore the potential of our method as an effective tool in advancing the field of geographic information systems and enhancing predictive analytics.